\documentclass[conference]{IEEEtran}
\IEEEoverridecommandlockouts
\usepackage{cite}
\usepackage{amsmath,amssymb,amsfonts}
\usepackage{graphicx}
\usepackage{textcomp}
\usepackage{xcolor}
\usepackage{subcaption}
\usepackage{siunitx}
\usepackage{float}

\def\BibTeX{{\rm B\kern-.05em{\sc i\kern-.025em b}\kern-.08em
    T\kern-.1667em\lower.7ex\hbox{E}\kern-.125emX}}

\begin{document}

\title{Robotic Budding: From Homogeneous Cells to Diverse Morphologies in Physical Self-Reproduction\\
\large\normalfont }

\author{Siyuan Zhang$^{1}$, Junyan Liu$^{1}$, Wanglin He$^{1}$, and Hod Lipson$^{1, \dagger}$
\thanks{*This work was supported by Creative Machines Lab in Columbia University.}%
\thanks{$^{1}$Siyuan Zhang, Junyan Liu, Wanglin He, and Hod Lipson are with the Department of Mechanical Engineering, Columbia University, New York, NY 10027, USA.}%
\thanks{$^{\dagger}${\raggedright Corresponding author: Hod Lipson ({hod.lipson@columbia.edu}).}}}%

\maketitle

\begin{abstract}
Physical self-reproduction in modular robots remains difficult because a robot must not only move, but also collect, align, dock, and separate modules reliably in hardware. 
Taking inspiration from how genetically identical cells form the varied tissues of an organism, we introduce a single standardized robotic module that combines a magnetic docking interface with two actuated joints and folds from a flat, two-dimensional configuration into three-dimensional structures.
Modules of this single design can be assembled into two morphologies: a three-module ``trilobite'' and a five-module ``seal''. 
Using a small set of discrete motion commands, a human operator steers an assembled parent robot to dock magnetically with a sequence of spare modules, which are assembled into a second, structurally identical individual that then separates from the parent.
Because the parent retains all of its original modules, this is a budding-style form of physical reproduction.
Here, we report preliminary hardware demonstrations of this full process on both morphologies.
We discuss the current scope of the system, including its present reliance on direct human teleoperation rather than autonomous control, and outline directions toward autonomous execution of this process and repeated validation across a larger number of independent trials.
\end{abstract}

\section{Introduction}

Modular robotic systems achieve versatility, robustness, and adaptability by composing a small set of standardized units into task-specific morphologies \cite{yim2007modular}. 
A particularly ambitious extension of this idea is physical self-reproduction: using the same modules that make up a robot to construct additional, independently functioning copies of it. 
This goal's theoretical foundations date back to von Neumann's theory of self-reproducing automata \cite{vonneumann1966theory}. 
Recent work has demonstrated kinematic self-replication in reconfigurable biological organisms \cite{kriegman2021kinematic}, but achieving comparable physical reproduction in purely electromechanical systems has remained largely limited to simulation or to structured, partially scaffolded environments \cite{zykov2005selfreproducing,zykov2007evolved,lee2007robotic}. 

\begin{figure*}
    \centering
    \includegraphics[width=0.8\linewidth]{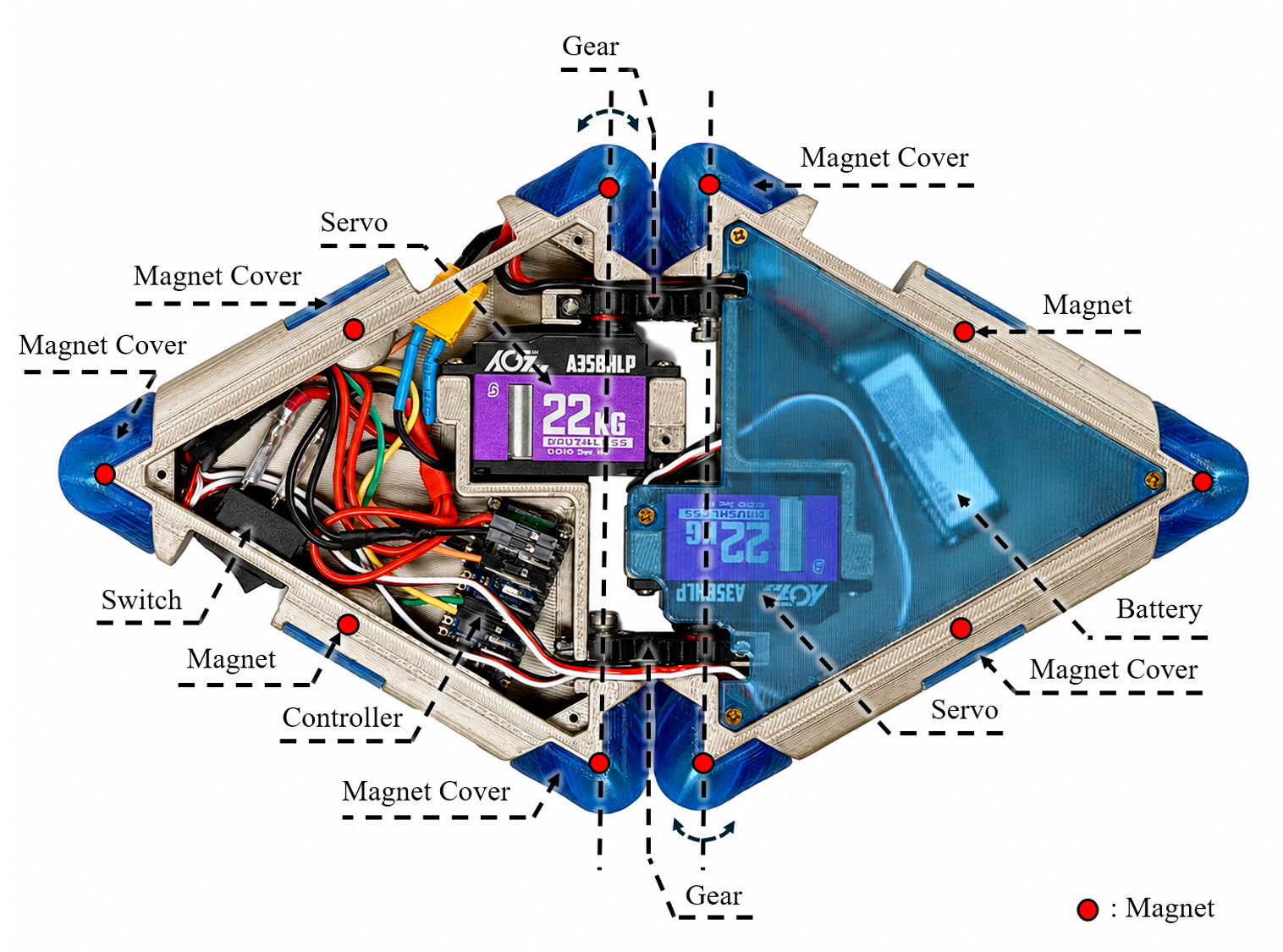}
    \caption{Internal structural layout and component allocation of an individual module, highlighting the dual-servo drive mechanism, onboard electronics, and magnetic docking interfaces (red dots).} 
    \label{fig:single_module_demo}
\end{figure*}

In this work, we investigate whether a single module design (Fig.~\ref{fig:single_module_demo}) can support both (i) structurally distinct target morphologies, and (ii) physical reproduction via module collection and assembly, without redesigning the hardware between cases. 
We report a system in which a human operator steers an assembled parent robot, through a small set of discrete motion commands relayed wirelessly to its modules, to dock with spare modules in sequence, culminating in the release of a structurally identical offspring.

To demonstrate reproduction across distinct morphologies, we implement two structurally and topologically distinct morphologies as proof of concept: a three-module “trilobite” featuring a segmented body with a short caudal extension, and a five-module “seal” equipped with a pair of rear flippers, as shown in Fig.~\ref{fig:morphologies}.
Together, they show that one module design can produce morphologies that differ in module count and connection topology. 
We demonstrate the full reproduction process on both morphologies.

The contribution of this work is a preliminary hardware demonstration and system description, rather than a comprehensive quantitative evaluation.

For a target morphology containing $n$ modules, let $\mathcal{O}=\{o_1,\ldots,o_n\}$ denote the modules initially composing the parent and let $\mathcal{E}=\{e_1,\ldots,e_n\}$ denote the modules acquired from the environment. The physical reproduction process considered here is therefore written as
\begin{equation}
R(\mathcal{O})+\mathcal{E}
\xrightarrow{\pi_{\mathrm{collect}}}
R(\mathcal{O}\cup\mathcal{E})
\xrightarrow{\pi_{\mathrm{separate}}}
R(\mathcal{O})+R(\mathcal{E}),
\label{eq:reproduction_cycle}
\end{equation}

where $R(\mathcal{S})$ denotes a connected robot formed from the module set $\mathcal{S}$, ``$+$'' denotes physically separate entities, $\pi_{\mathrm{collect}}$ denotes the teleoperated collection of spare modules
followed by their assembly and folding into the three-dimensional offspring body, and $\pi_{\mathrm{separate}}$ denotes the pre-programmed separation sequence.
Let $G_n^{\star}$ denote the target connection graph of an $n$-module morphology, and let $G_{\mathrm{p}}$ and $G_{\mathrm{o}}$ denote the connection graphs of the separated parent and offspring. 
Structural reproduction requires
\begin{equation}
G_{\mathrm{p}}\simeq G_{\mathrm{o}}\simeq G_n^{\star},
\qquad
|V(G_{\mathrm{p}})|=|V(G_{\mathrm{o}})|=n,
\label{eq:structural_equivalence}
\end{equation}
where $\simeq$ denotes graph isomorphism. 
Condition (2) is necessary but not sufficient, because a connection graph does not encode joint configuration: the flat precursor and the folded offspring share the same graph (Fig. 4, VI and VII). 
To make the material lineage explicit, we define the module-provenance segregation index
\begin{equation}
\sigma=
\frac{|V(G_{\mathrm{p}})\cap\mathcal{O}|+
      |V(G_{\mathrm{o}})\cap\mathcal{E}|}{2n}.
\label{eq:provenance}
\end{equation}
The demonstrated process has $\sigma=1$: all original modules remain in the parent, whereas all acquired modules constitute the offspring. 
Thus, although the transient $2n$-module composite separates into two equal-sized $n$-module robots, the reproduction mode is more accurately described as equal-sized budding-style reproduction rather than mitosis or binary fission.

\section{System Overview}
\label{sec:system_overview}

As shown in Fig.~\ref{fig:single_module_demo}, each module is composed of two rigid triangular segments connected in series by two revolute joints. Each joint is driven by an internal servo actuator and position-controlled across approximately $90^{\circ}$, providing a total relative articulation range of $\pm180^{\circ}$ between the two segments. This range of motion allows the module to fold from a flat, planar configuration into three-dimensional structures. Internally, the hollow triangular frames house a $7.4\text{ V}$ Li-Po battery power unit and an onboard Arduino Nano ESP32 microcontroller, with the two central $22\text{ kgf}\cdot\text{cm}$ ($\approx 2.16\text{ N}\cdot\text{m}$) servos synchronized via integrated gears mechanism. For inter-module coupling, each segment terminates in a triangular end equipped with five N42 neodymium magnets embedded inside protective covers: three situated at the vertices of the equilateral triangle, and two at the midpoints of its outward-facing edges. Because either end of a module can dock with any end of another module via these connecting edges, the hardware design imposes no directional pairing constraints.

\begin{figure*}[t]
\centering
\begin{subfigure}[b]{0.32\textwidth}
\centering
\includegraphics[width=\linewidth]{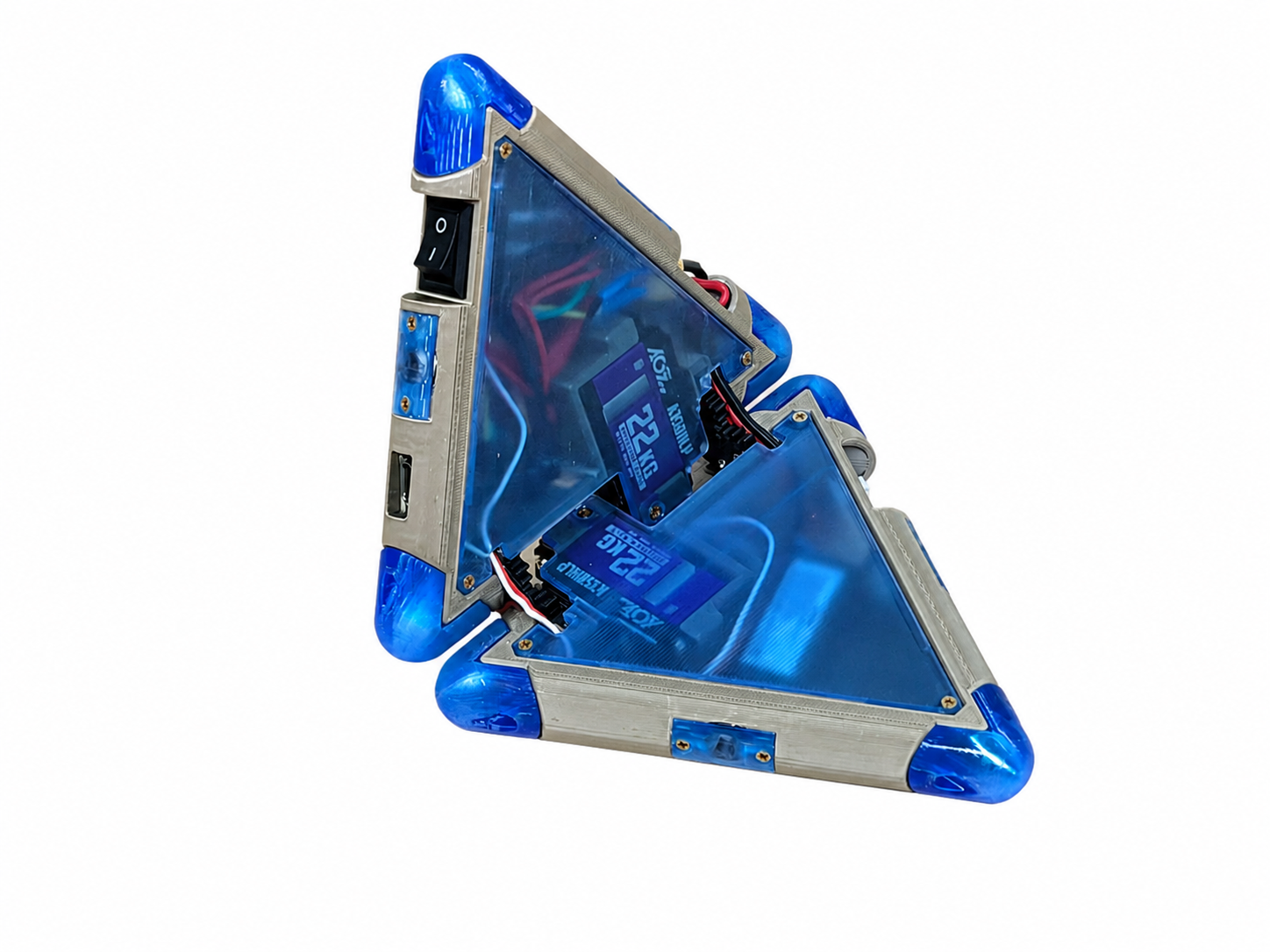}
\caption{Single module, folded $90^{\circ}$, close-up view}
\label{fig:module_folded}
\end{subfigure}
\hfill
\begin{subfigure}[b]{0.32\textwidth}
\centering
\includegraphics[width=\linewidth]{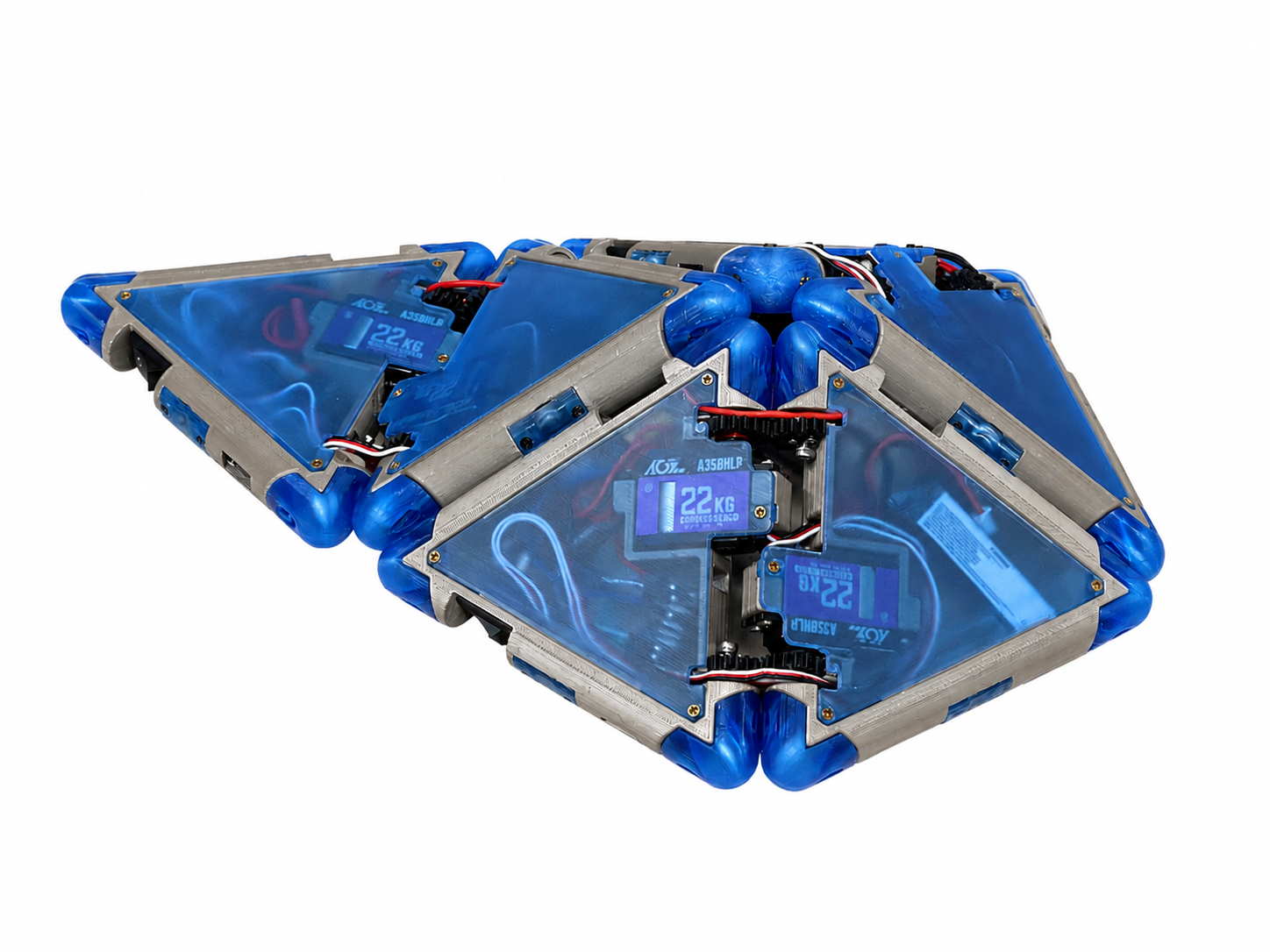}
\caption{Trilobite, close-up view}
\label{fig:trilobite_oblique}
\end{subfigure}
\hfill
\begin{subfigure}[b]{0.32\textwidth}
\centering
\includegraphics[width=\linewidth]{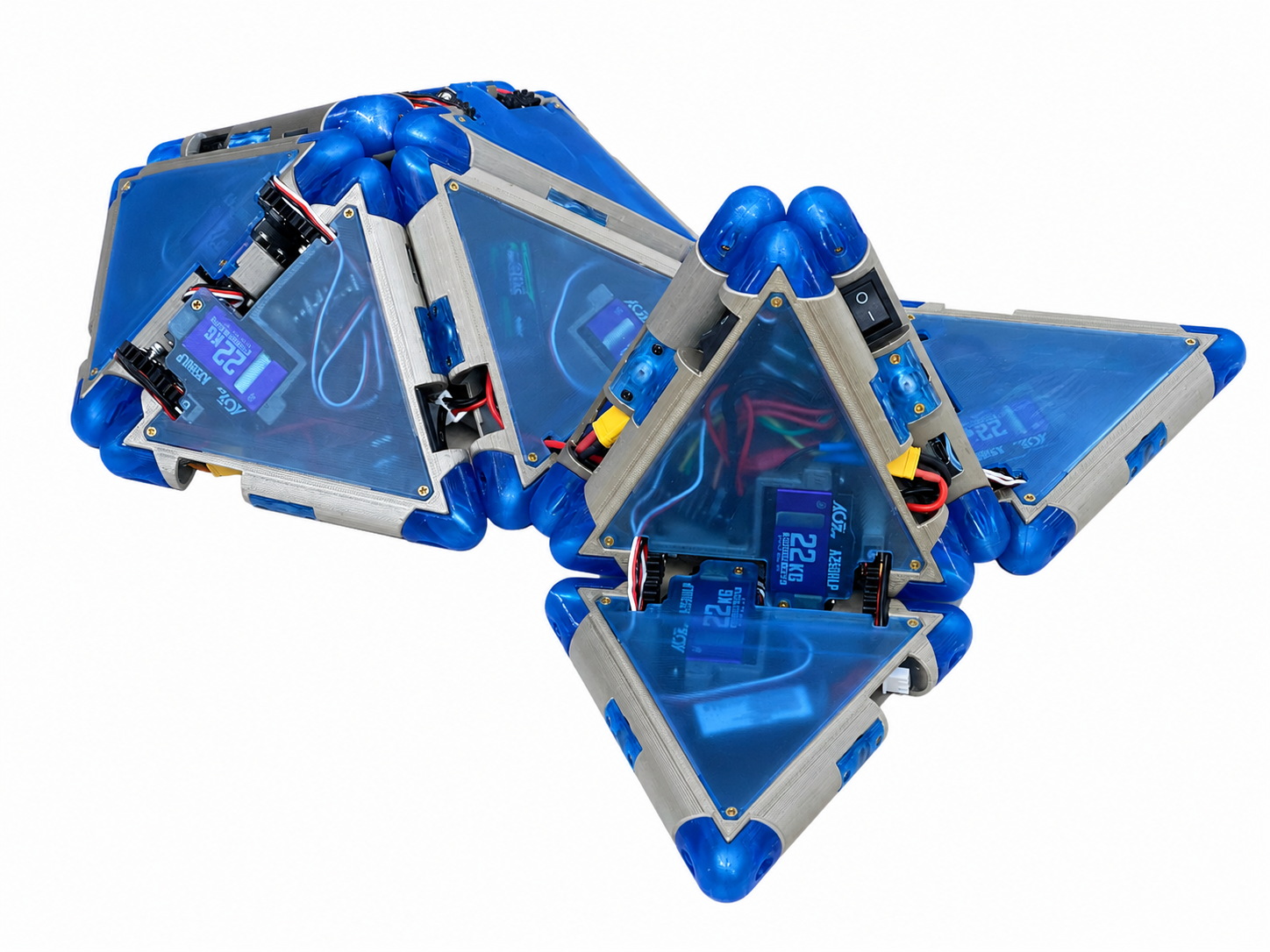}
\caption{Seal, close-up view}
\label{fig:seal_oblique}
\end{subfigure}
\caption{(a) A single module folded
to $90^{\circ}$ between its two segments, illustrating the joint range described in
Section~\ref{sec:system_overview}. (b) The assembled three-module trilobite morphology. (c) The assembled five-module
seal morphology.}
\label{fig:morphologies}
\end{figure*}

Two target morphologies are assembled from this single module design: a three-module ``trilobite''
and a five-module ``seal'' (Fig.~\ref{fig:morphologies}). Motion of an assembled morphology,
whether as a stand-alone locomoting robot or as a parent robot performing module retrieval, is
composed from three discrete motion primitives: a forward step, a left turn, and a right turn.
These are executed by microcontroller firmware in response to commands issued over a serial link.

The module design relies on a magnetic docking interaction: an attractive force that falls off
with the fourth power of separation distance, combined with velocity-dependent damping near
contact. That study validated the numerical stability and reliable
attachment behavior of the chosen docking parameters across a range of separation distances,
independent of any specific target morphology.

\section{Teleoperated Collection and Assembly}

Collection and assembly are currently carried out by direct human teleoperation rather than
autonomous control. An operator observes the parent robot and the scattered spare modules
directly, and one at a time issues a discrete motion command, a forward step, a left turn, or a
right turn, through a browser-based control terminal. The terminal is connected over a USB
serial link to a host microcontroller, which wirelessly broadcasts the corresponding command to
the target modules; each module in turn sends a periodic heartbeat signal back to the host over
the same wireless link to report its status. Guided
by this real-time visual feedback, the operator steers the parent toward each target module in
turn until its magnetic contact points engage. Docking is judged by the operator based on direct
observation, without an independent physical confirmation, such as force or current sensing,
that a magnetic bond has actually formed.

\begin{figure}[t]
\centering
\includegraphics[width=\linewidth]{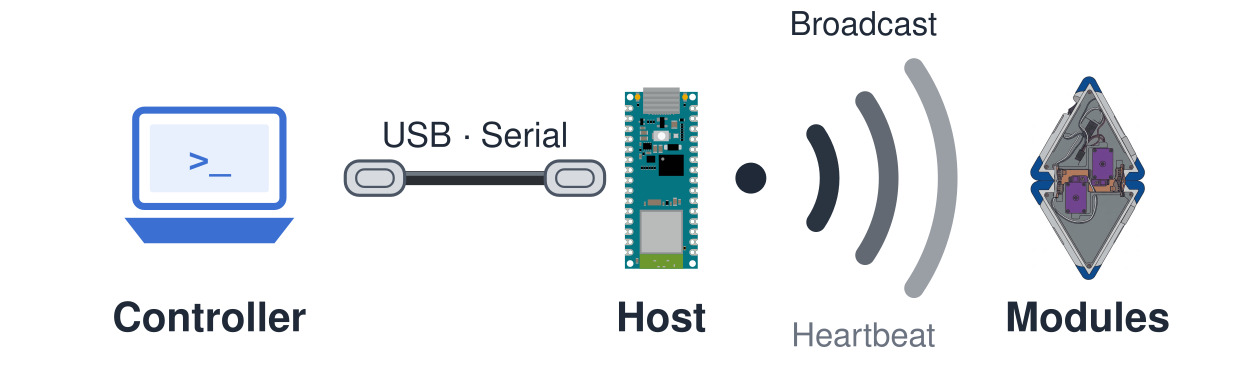}
\caption{Control system communication architecture. The operator issues commands from a
browser-based control terminal on a laptop (Controller), connected over a USB serial link to a
host microcontroller (Host). The Host wirelessly broadcasts each command to the modules, and each
module sends a periodic heartbeat signal back to the Host over the same wireless link to report
its status.}
\label{fig:control_diagram}
\end{figure}

Once all modules for a given morphology have been collected, the operator triggers a
pre-programmed separation sequence: distinct, timed motion paths for the parent and the newly
formed offspring, executed as a single firmware-level command. The offspring is then free to
move independently, and the operator issues several additional turn commands to it to demonstrate
this independent locomotion.

Let $C$ denote the set of magnetic connections intended to open during separation. The desired
topological result is
\begin{equation}
G_{2n}\setminus C=G_n^{(\mathrm{p})}\cup G_n^{(\mathrm{o})},
\qquad
G_n^{(\mathrm{p})}\cap G_n^{(\mathrm{o})}=\varnothing.
\label{eq:graph_cut}
\end{equation}
Separation is achieved by antagonistic motions of the connected modules, which generate
sufficient relative displacement and separating force to disengage the target magnetic interface
while preserving the remaining connections. In the present implementation, satisfaction of this
criterion is assessed from direct observation of the demonstration rather than by automated
analysis of the connection graph.

For each trial, we record the number of successful dockings, the total collection time, whether
separation succeeded, and whether the offspring could execute post-separation locomotion
commands.

\section{Preliminary Results}

\begin{figure}
\centering
\captionsetup[subfigure]{font=scriptsize, skip=0pt, labelformat=empty}

\begin{subfigure}[t]{0.49\columnwidth}
\centering
\includegraphics[width=\linewidth]{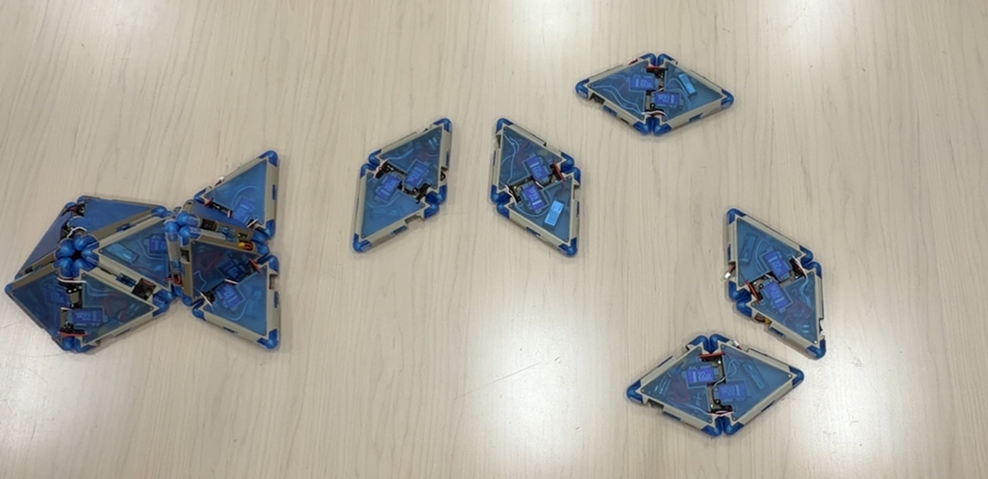}
\caption{Stage I: initial state (5 spare modules).}
\label{fig:seal_s1}
\end{subfigure}
\hfill
\begin{subfigure}[t]{0.49\columnwidth}
\centering
\includegraphics[width=\linewidth]{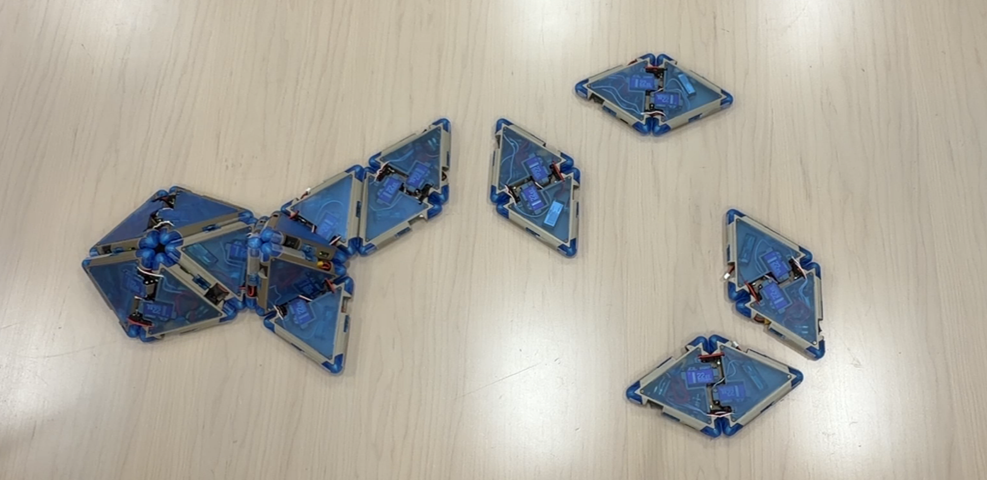}
\caption{Stage II: 1st module acquired.}
\label{fig:seal_s2}
\end{subfigure}

\vspace{1pt}

\begin{subfigure}[t]{0.49\columnwidth}
\centering
\includegraphics[width=\linewidth]{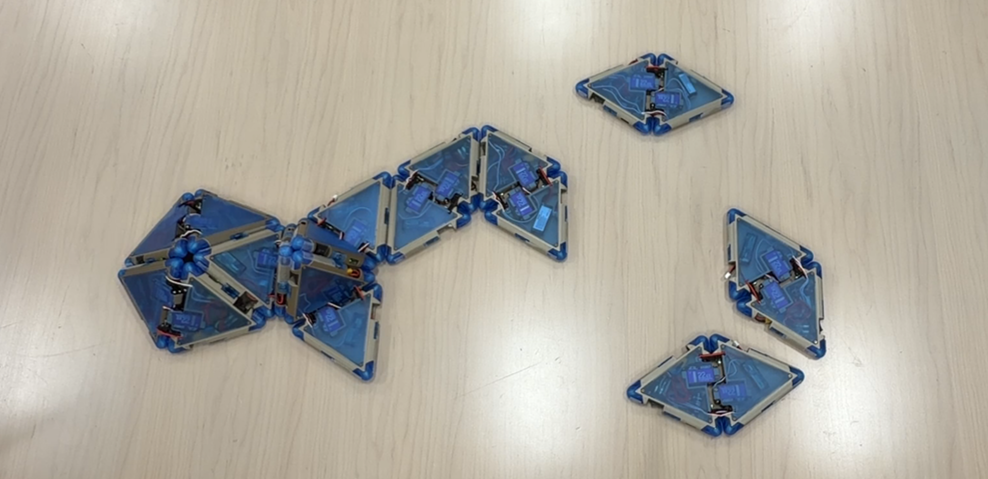}
\caption{Stage III: 2nd module acquired.}
\label{fig:seal_s3}
\end{subfigure}
\hfill
\begin{subfigure}[t]{0.49\columnwidth}
\centering
\includegraphics[width=\linewidth]{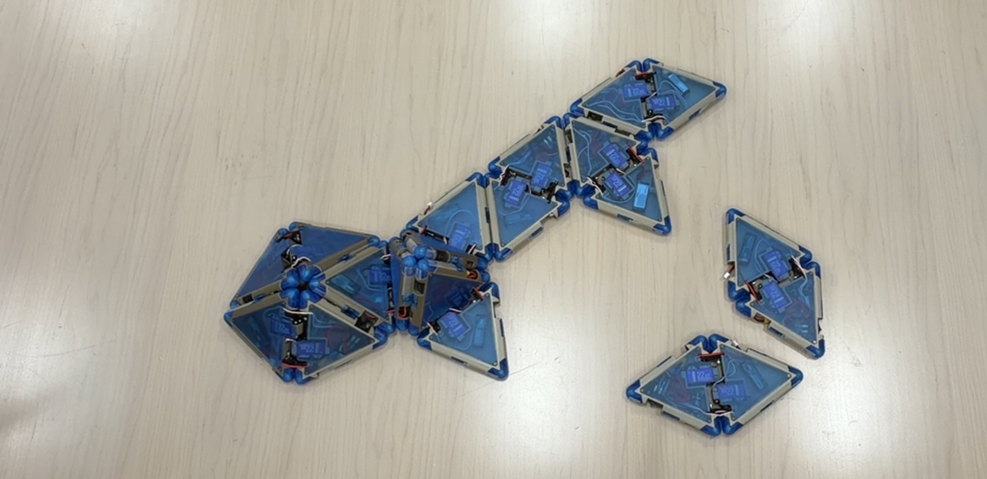}
\caption{Stage IV: 3rd module acquired.}
\label{fig:seal_s4}
\end{subfigure}

\vspace{1pt}

\begin{subfigure}[t]{0.49\columnwidth}
\centering
\includegraphics[width=\linewidth]{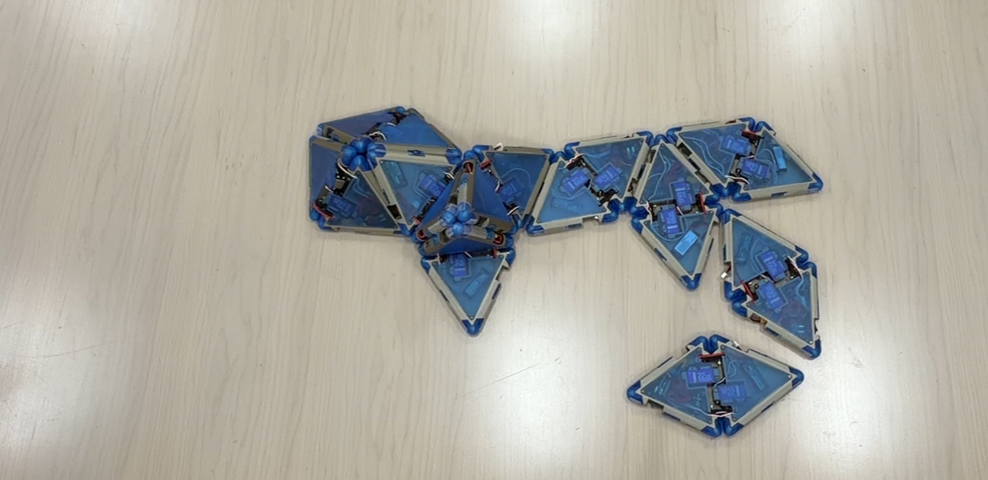}
\caption{Stage V: 4th module acquired.}
\label{fig:seal_s5}
\end{subfigure}
\hfill
\begin{subfigure}[t]{0.49\columnwidth}
\centering
\includegraphics[width=\linewidth]{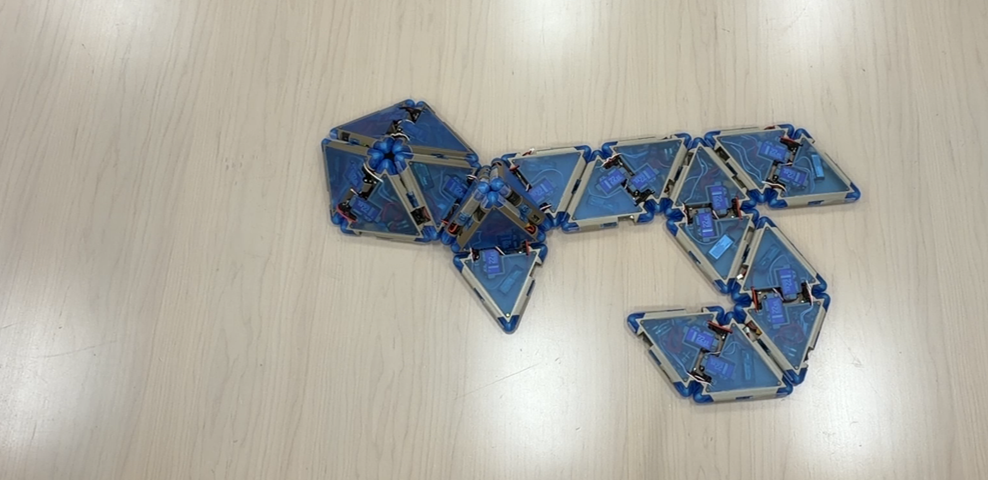}
\caption{Stage VI: 5th module acquired (2-D).}
\label{fig:seal_s6}
\end{subfigure}

\vspace{1pt}

\begin{subfigure}[t]{0.49\columnwidth}
\centering
\includegraphics[width=\linewidth]{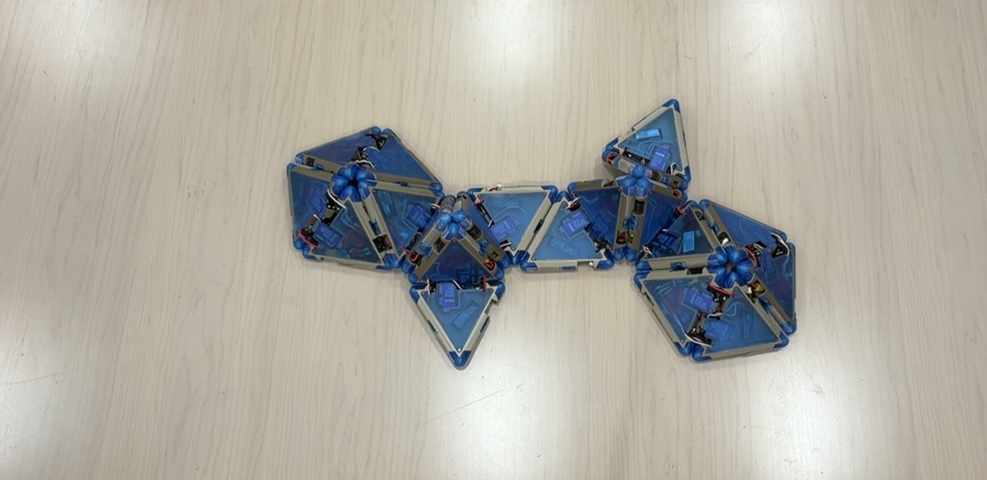}
\caption{Stage VII: 3-D offspring folding.}
\label{fig:seal_s7}
\end{subfigure}
\hfill
\begin{subfigure}[t]{0.49\columnwidth}
\centering
\includegraphics[width=\linewidth]{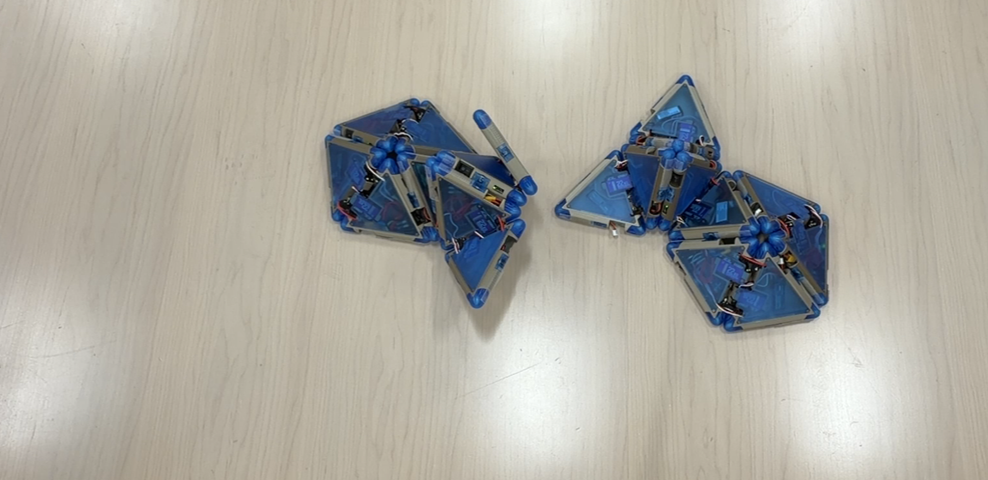}
\caption{Stage VIII: parent-offspring separation.}
\label{fig:seal_s8}
\end{subfigure}

\caption{Eight stages of the five-module seal reproduction process, ordered from left to right and top to bottom. The parent sequentially acquires five initially loose environmental modules (I--VI). The acquired modules then fold from two-dimensional configurations into a three-dimensional offspring while still connected to the parent (VII). Selective magnetic separation produces two independent five-module seal robots (VIII): the persistent parent contains the five original modules, and the offspring contains the five acquired modules.}
\label{fig:seal_reproduction_sequence}
\end{figure}

We demonstrate the full collection-and-assembly process on the five-module seal morphology: an
assembled parent seal, under direct human teleoperation, located and magnetically docked with
five scattered spare modules in sequence. The parent then executed the separation sequence, and
a structurally identical offspring seal separated from the unchanged parent and was subsequently
driven to execute several turns, demonstrating independent locomotion equivalent to that of the
parent. Figure.~\ref{fig:seal_reproduction_sequence} records eight stages of this complete reproduction
process. In the initial state (I), the parent is a three-dimensional seal composed of five
modules, and five additional modules are distributed separately in the environment. Sequential
collection is shown in II--V: the parent has acquired one, two, three, and four modules,
respectively, while the number of loose environmental modules decreases from four to one. After
the fifth acquisition (VI), the composite system contains $2n=10$ modules; all five acquired
modules are connected to the parent but remain in their flat, two-dimensional configurations.
The acquired modules subsequently fold into the three-dimensional seal morphology while the two
five-module bodies remain connected (VII), producing the pre-separation configuration. Finally,
the programmed antagonistic motion releases the target magnetic interface and produces a fully
separated five-module parent and five-module offspring, both in their three-dimensional seal
morphologies (VIII). The original five modules remain in the parent and the five acquired modules
constitute the offspring, consistent with the provenance condition $\sigma=1$. Because the two
separated individuals share the same morphology, actuation, and docking hardware, each is
physically equipped for subsequent module collection and separation, although recursive
reproduction by the offspring has not yet been experimentally tested.

\begin{figure}
\centering
\captionsetup[subfigure]{font=scriptsize, skip=0pt, labelformat=empty}
\begin{subfigure}[t]{0.49\columnwidth}
\centering
\includegraphics[width=\linewidth]{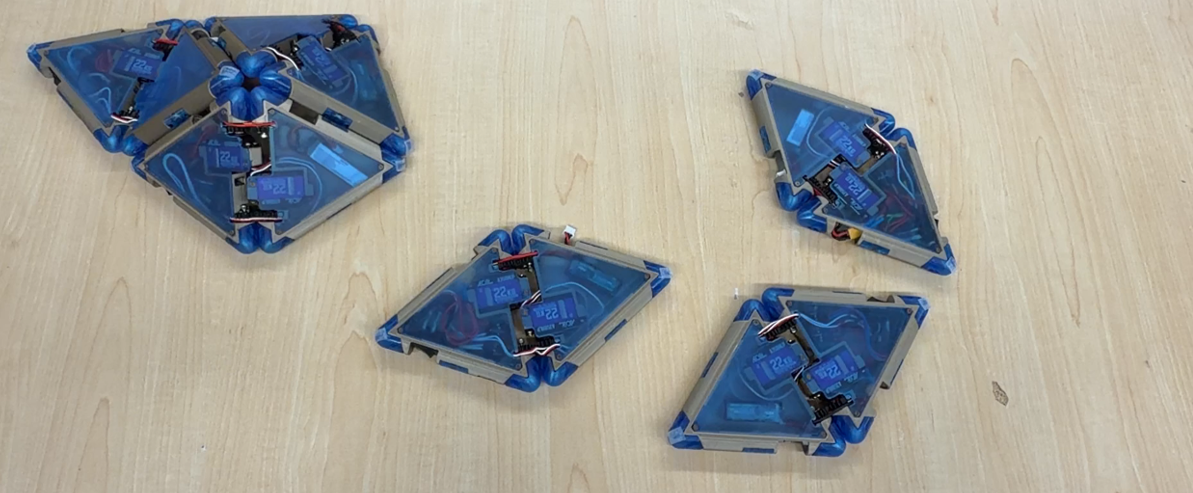}
\caption{Stage I: initial state (3 spare modules).}
\label{fig:trilobite_b1}
\end{subfigure}
\hfill
\begin{subfigure}[t]{0.49\columnwidth}
\centering
\includegraphics[width=\linewidth]{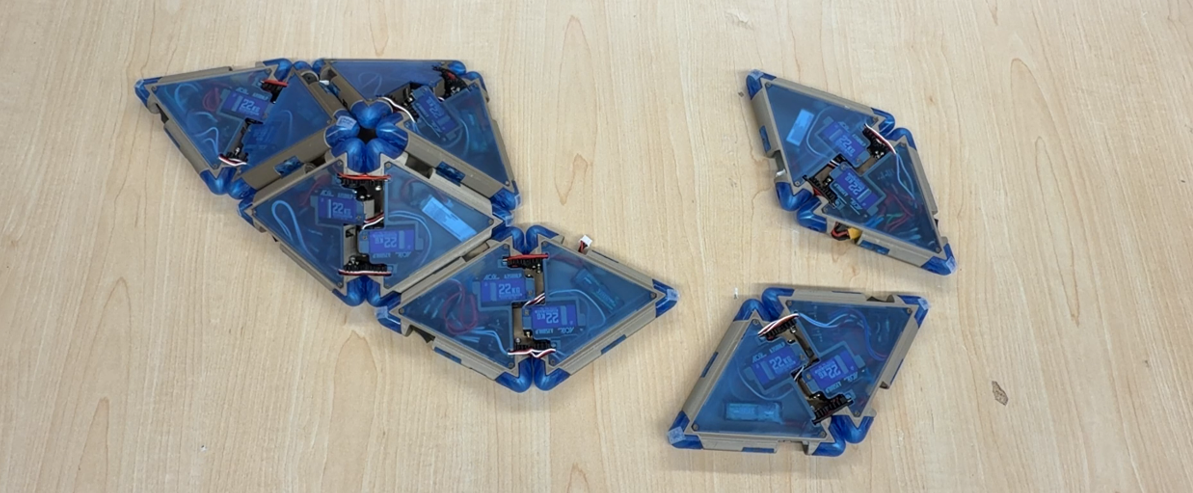}
\caption{Stage II: 1st module acquired.}
\label{fig:trilobite_b2}
\end{subfigure}

\vspace{1pt}

\begin{subfigure}[t]{0.49\columnwidth}
\centering
\includegraphics[width=\linewidth]{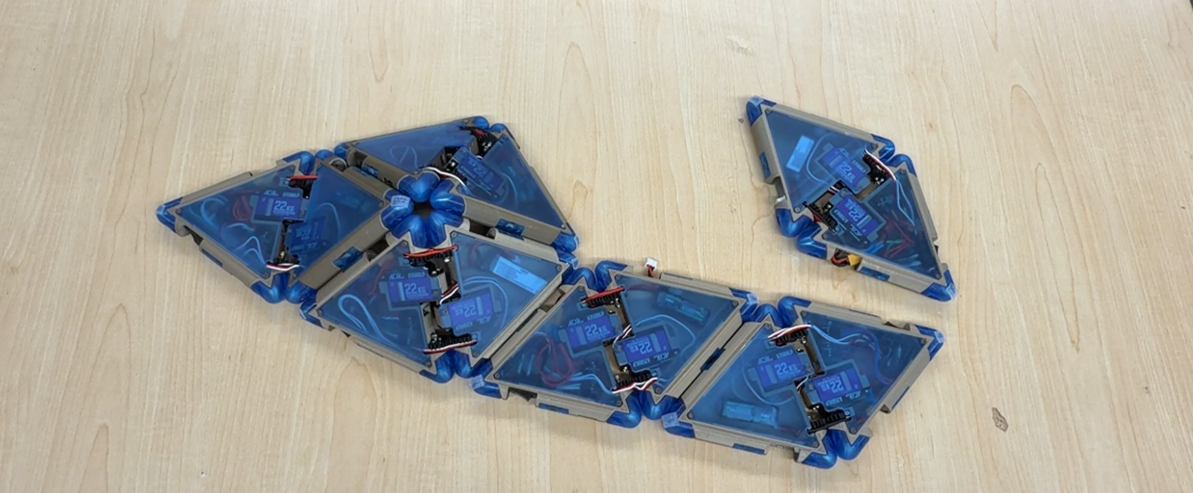}
\caption{Stage III: 2nd module acquired.}
\label{fig:trilobite_b3}
\end{subfigure}
\hfill
\begin{subfigure}[t]{0.49\columnwidth}
\centering
\includegraphics[width=\linewidth]{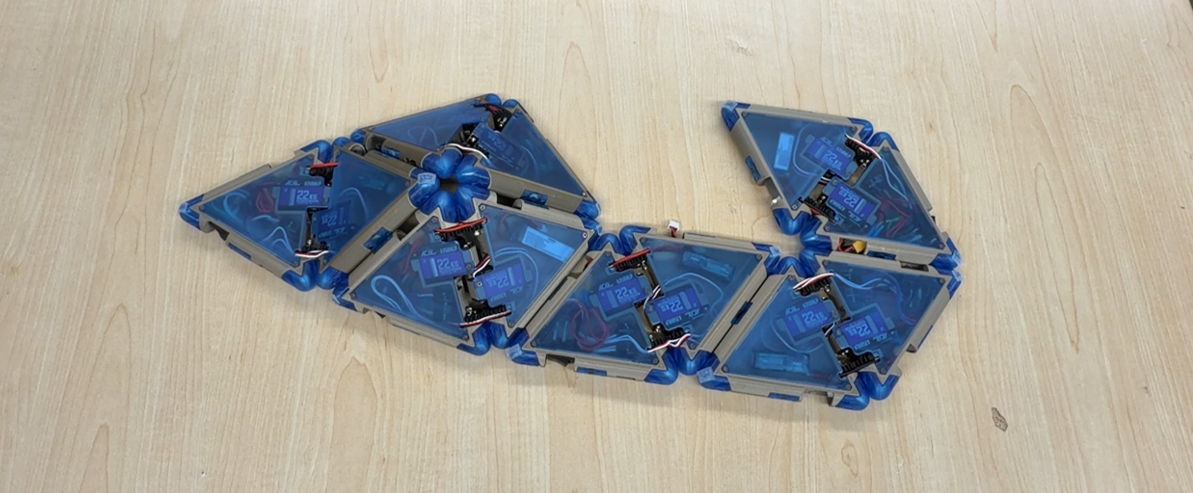}
\caption{Stage IV: 3rd module acquired (2-D).}
\label{fig:trilobite_b4}
\end{subfigure}

\vspace{1pt}

\begin{subfigure}[t]{0.49\columnwidth}
\centering
\includegraphics[width=\linewidth]{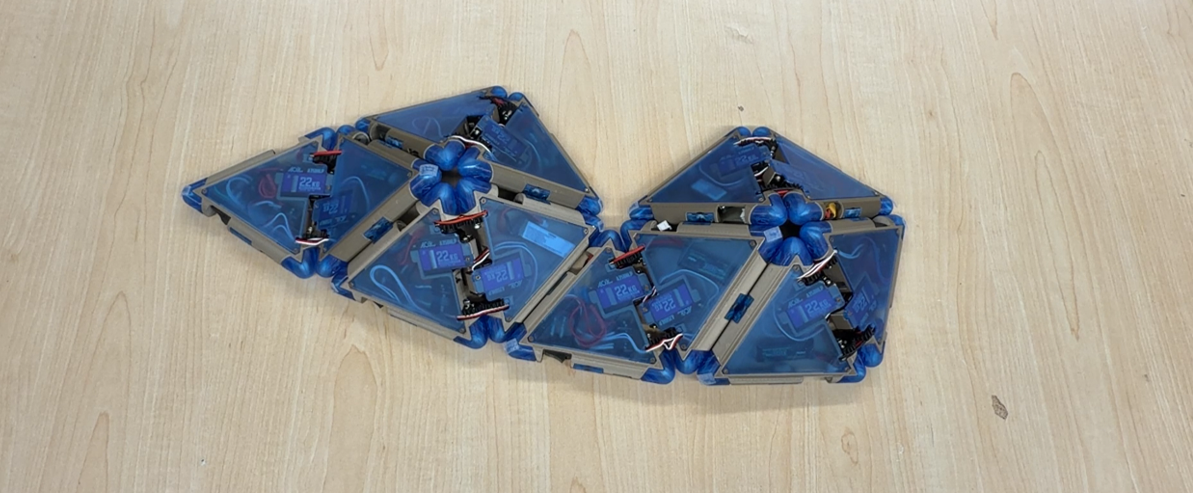}
\caption{Stage V: 3-D offspring folding.}
\label{fig:trilobite_b5}
\end{subfigure}
\hfill
\begin{subfigure}[t]{0.49\columnwidth}
\centering
\includegraphics[width=\linewidth]{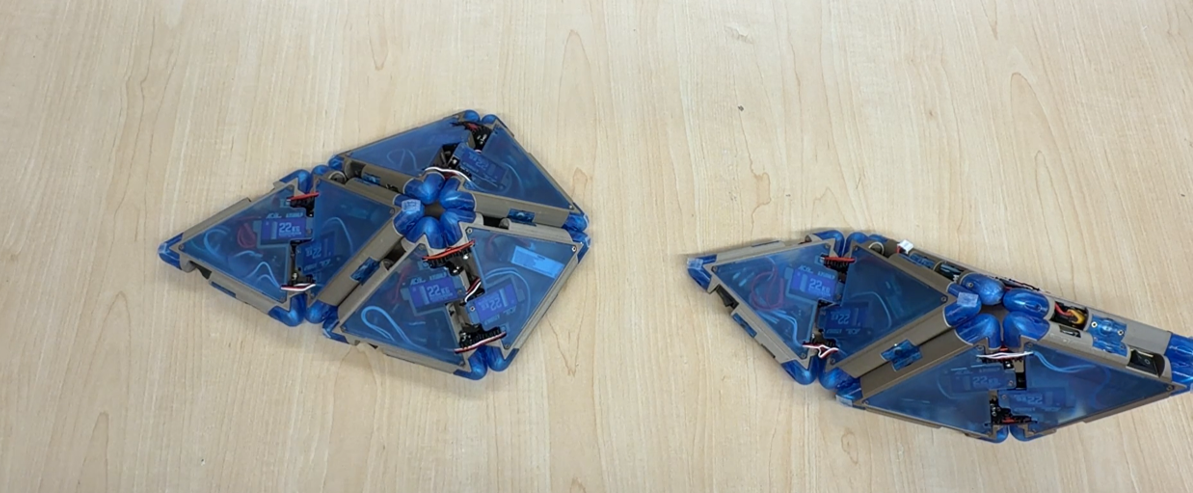}
\caption{Stage VI: parent-offspring separation.}
\label{fig:trilobite_b6}
\end{subfigure}
\caption{Six stages of the three-module trilobite reproduction process, ordered from left to right
and top to bottom. The parent sequentially acquires three initially loose environmental modules
(I--IV). The acquired modules then fold from two-dimensional configurations into a
three-dimensional offspring while still connected to the parent (V). Selective magnetic
separation produces two independent three-module trilobite robots (VI): the persistent parent contains
the three original modules, and the offspring contains the three acquired modules.}
\label{fig:trilobite_reproduction_sequence}
\end{figure}

As shown in Figure.~\ref{fig:trilobite_reproduction_sequence}. The three-module trilobite morphology was assembled from the same module design, illustrating
that a single module supports both target morphologies. The full collection-and-reproduction
process was also demonstrated on the trilobite morphology: an assembled parent trilobite, under
direct human teleoperation, located and magnetically docked with three scattered spare modules
in sequence, executed the separation sequence, and produced a structurally identical offspring.
The results reported here are single successful demonstrations on each morphology rather than
statistically characterized trial series; a more complete quantitative evaluation is in
preparation.

\section{Discussion and Future Work}

Here, we show that a single module combining a magnetic docking interface with two actuated joints can support multiple target morphologies and physical, budding-style reproduction in an unstructured workspace \cite{zykov2005selfreproducing,zykov2007evolved,lee2007robotic}. Although module navigation and docking were teleoperated rather than autonomous, the demonstration confirms that the design mechanically supports the full end-to-end assembly and separation process.

The current implementation has several limitations that bound the scope of these results. Most
significantly, collection and assembly are currently teleoperated rather than autonomous:
navigation and docking decisions are made by a human operator based on direct observation, rather
than by an onboard or vision-based controller. Docking success is judged by the operator from
direct observation alone, without independent physical confirmation. The results reported above
are single successful demonstrations on each morphology, rather than a statistically meaningful
trial series.
Finally, while the offspring produced is structurally identical to its parent and capable of
independent locomotion, whether it can itself repeat the collection-and-assembly process to
produce a further generation has not yet been tested.

Ongoing work targets these limitations directly: replacing teleoperation with autonomous,
sensor-guided control of the collection-and-assembly process; repeating the demonstration across
a larger number of independent trials on both morphologies; and testing whether an
offspring can itself produce a subsequent generation. A more complete study incorporating these
results is in preparation.

\end{document}